\documentclass{article}
\usepackage[utf8]{inputenc}
\usepackage{main}
\usepackage{microtype}
\usepackage{subcaption}
\usepackage{graphicx}
\usepackage{times}
\usepackage{latexsym}
\usepackage{amsfonts}
\usepackage{subcaption}
\usepackage{amsmath}
\usepackage[table]{xcolor}
\usepackage{float}
\usepackage{footnote}
\usepackage{enumitem}
\usepackage{makecell}
\usepackage{bm}
\usepackage{arydshln}
\usepackage{pifont}
\usepackage{booktabs}
\usepackage{amssymb}
\usepackage{multicol}
\usepackage{multirow}
\usepackage{color}  
\usepackage{colortbl}
\usepackage{bbding}
\usepackage{makecell}
\usepackage{mathtools}
\usepackage{imakeidx}
\usepackage{longtable}
\usepackage{wrapfig}
\usepackage{mathrsfs}
\makeindex
\usepackage{arydshln}
\usepackage{lipsum}
\usepackage{natbib}
\usepackage[toc]{multitoc}
\usepackage[edges]{forest}
\usepackage[normalem]{ulem}
\definecolor{mydarkblue}{rgb}{0,0.08,0.45}
\usepackage[colorlinks=true,linkcolor=mydarkblue,citecolor=mydarkblue,filecolor=mydarkblue,urlcolor=mydarkblue]{hyperref}
\usepackage{cleveref}
\usepackage{CJKutf8}
\usepackage{awesomebox} 
\usepackage{bbding}
\usepackage[most]{tcolorbox}
\usepackage{booktabs}
\usepackage{geometry}
\definecolor{wkblue}{rgb}{0.2, 0.3, 0.6}
\definecolor{meta-color}{rgb}{0.5, 0.5, 0.5}
\usepackage{amsmath}
\usepackage{enumitem}

\usepackage[tikz]{bclogo}
\usepackage[framemethod=tikz]{mdframed}
\definecolor{bgblue}{RGB}{245,243,253}
\definecolor{ttblue}{RGB}{91,194,224}

\mdfdefinestyle{mystyle}{%
  rightline=true,
  innerleftmargin=10,
  innerrightmargin=10,
  outerlinewidth=3pt,
  topline=false,
  rightline=true,
  bottomline=false,
  skipabove=\topsep,
  skipbelow=\topsep
}

\newtcolorbox{myboxi}[1][]{
  breakable,
  title=#1,
  colback=red!5,
  colbacktitle=red!5,
  coltitle=black,
  fonttitle=\bfseries,
  bottomrule=0pt,
  toprule=0pt,
  leftrule=2pt,
  rightrule=2pt,
  titlerule=0pt,
  arc=0pt,
  outer arc=0pt,
  colframe=red,
}

\newtcolorbox{myboxnote}[1][]{
  breakable,
  title=#1,
  colback=orange!0,
  colbacktitle=orange!0,
  coltitle=black,
  fonttitle=\bfseries,
  bottomrule=0pt,
  toprule=0pt,
  leftrule=2pt,
  rightrule=2pt,
  titlerule=0pt,
  arc=0pt,
  outer arc=0pt,
  colframe=orange,
}

\newtcolorbox{myboxii}[1][]{
  breakable,
  freelance,
  title=#1,
  colback=white,
  colbacktitle=white,
  coltitle=black,
  fonttitle=\bfseries,
  bottomrule=0pt,
  boxrule=0pt,
  colframe=white,
  overlay unbroken and first={
  \draw[red!75!black,line width=3pt]
    ([xshift=5pt]frame.north west) -- 
    (frame.north west) -- 
    (frame.south west);
  \draw[red!75!black,line width=3pt]
    ([xshift=-5pt]frame.north east) -- 
    (frame.north east) -- 
    (frame.south east);
  },
  overlay unbroken app={
  \draw[red!75!black,line width=3pt,line cap=rect]
    (frame.south west) -- 
    ([xshift=5pt]frame.south west);
  \draw[red!75!black,line width=3pt,line cap=rect]
    (frame.south east) -- 
    ([xshift=-5pt]frame.south east);
  },
  overlay middle and last={
  \draw[red!75!black,line width=3pt]
    (frame.north west) -- 
    (frame.south west);
  \draw[red!75!black,line width=3pt]
    (frame.north east) -- 
    (frame.south east);
  },
  overlay last app={
  \draw[red!75!black,line width=3pt,line cap=rect]
    (frame.south west) --
    ([xshift=5pt]frame.south west);
  \draw[red!75!black,line width=3pt,line cap=rect]
    (frame.south east) --
    ([xshift=-5pt]frame.south east);
  },
}

\usepackage{fancyhdr} 
\usepackage{blindtext} 

\DeclareCaptionFont{black}{\color{black}}

\definecolor{myblue}{rgb}{0.9, 0.1, 0.94}
\definecolor{mygreen}{rgb}{0.64, 0.56, 0.88}
\definecolor{myyellow}{rgb}{0.68, 0.6, 0.1}
\definecolor{fancygreen}{rgb}{0.33, 0.68, 0.20}
\definecolor{salmon}{rgb}{0.94, 0.52, 0.49}
\definecolor{tablegreen}{rgb}{0.82, 0.94, 0.75}
\definecolor{tableblue}{rgb}{0.81, 0.90, 0.94}
\definecolor{tablered}{rgb}{0.97, 0.85, 0.85}
\definecolor{tableorange}{rgb}{0.96, 0.85, 0.81}
\definecolor{TiffanyBlue}{rgb}{0.0, 0.68, 0.67}

\newenvironment{itemize*}%
 {\leftmargini=10pt\begin{itemize}%
  \setlength{\itemsep}{0pt}%
  \setlength{\parskip}{0pt}%
  }%
 {\end{itemize}}
\newenvironment{enumerate*}%
 {\begin{enumerate}%
  \setlength{\itemsep}{0pt}%
  \setlength{\parskip}{0pt}}%
 {\end{enumerate}}

\usepackage{listings}

\newcommand\JSONnumbervaluestyle{\color{blue}}
\newcommand\JSONstringvaluestyle{\color{red}}

\newif\ifcolonfoundonthisline

\makeatletter

\lstdefinestyle{json}
{
  showstringspaces    = false,
  keywords            = {false,true},
  alsoletter          = 0123456789.,
  morestring          = [s]{"}{"},
  stringstyle         = \ifcolonfoundonthisline\JSONstringvaluestyle\fi,
  MoreSelectCharTable =%
    \lst@DefSaveDef{`:}\colon@json{\processColon@json},
  basicstyle          = \ttfamily,
  keywordstyle        = \ttfamily\bfseries,
}

\newcommand\processColon@json{%
  \colon@json%
  \ifnum\lst@mode=\lst@Pmode%
    \global\colonfoundonthislinetrue%
  \fi
}

\lst@AddToHook{Output}{%
  \ifcolonfoundonthisline%
    \ifnum\lst@mode=\lst@Pmode%
      \def\lst@thestyle{\JSONnumbervaluestyle}%
    \fi
  \fi
  \lsthk@DetectKeywords%
}

\lst@AddToHook{EOL}%
  {\global\colonfoundonthislinefalse}

\makeatother

\usepackage{etoolbox}
\usepackage{natbib}
\usepackage{url}
\newcounter{bibcount}
\makeatletter
\patchcmd{\@lbibitem}{\item[}{\item[\hfil\stepcounter{bibcount}{[\thebibcount]}}{}{}
\renewcommand\NAT@bibsetup%
  [1]{\setlength{\leftmargin}{\bibhang}\setlength{\itemindent}{-\parindent}%
      \setlength{\itemsep}{\bibsep}\setlength{\parsep}{\z@}}
\makeatother

\newcommand*\samethanks[1][\value{footnote}]{\footnotemark[#1]}

\author{
Yakun Zhu$^{1,3}$\thanks{~~Co-first authors}\space\space\space\space
Zhongzhen Huang$^{1,3}$\samethanks\space\space\space\space
Linjie Mu$^{1,3}$\samethanks\space\space\space\space Yutong Huang$^{1}$\\
\textbf{Wei Nie$^{5}$\space\space\space
Jiaji Liu$^{6}$\space\space\space
Shaoting Zhang$^{1}$\thanks{~~Corresponding author}\space\space\space
Pengfei Liu$^{1, 2, 4}$\samethanks\space\space\space
Xiaofan Zhang$^{1,2,3}$\samethanks}\\
$^1$Shanghai Jiao Tong University , $^2$SII, $^3$SPIRAL Lab, $^4$Generative AI Research Lab (GAIR)\\ $^5$Shanghai Chest Hospital, $^6$Beijing Anzhen Hospital, Capital Medical University
}

\begin{document}

\title{DiagnosisArena: Benchmarking Diagnostic Reasoning\\ for Large Language Models}

\maketitle
\thispagestyle{fancy}
\fancyhead{}
\lhead{}
\lhead{\includegraphics[height=0.67cm]{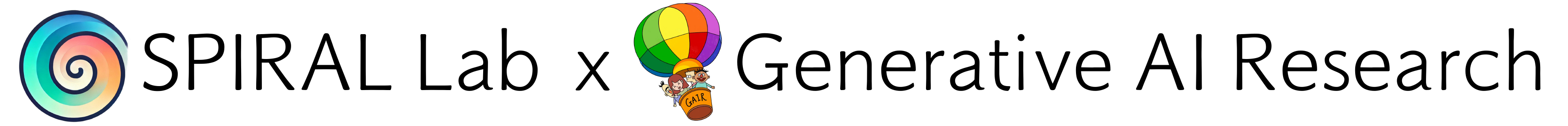}}
\renewcommand{\headrulewidth}{0pt}
\setlength{\headsep}{0mm}

\begin{abstract}
The emergence of groundbreaking large language models capable of performing complex reasoning tasks holds significant promise for addressing various scientific challenges, including those arising in complex clinical scenarios. To enable their safe and effective deployment in real-world healthcare settings, it is urgently necessary to benchmark the diagnostic capabilities of current models systematically.
Given the limitations of existing medical benchmarks in evaluating advanced diagnostic reasoning, we present \textit{DiagnosisArena}, a comprehensive and challenging benchmark designed to rigorously assess professional-level diagnostic competence. \textit{DiagnosisArena} consists of 1,113 pairs of segmented patient cases and corresponding diagnoses, spanning 28 medical specialties, deriving from clinical case reports published in 10 top-tier medical journals. The benchmark is developed through a meticulous construction pipeline, involving multiple rounds of screening and review by both AI systems and human experts, with thorough checks conducted to prevent data leakage.
Our study reveals that even the most advanced reasoning models, o3, o1, and DeepSeek-R1, achieve only 51.12\%, 31.09\%, and 17.79\% accuracy, respectively. This finding highlights a significant generalization bottleneck in current large language models when faced with clinical diagnostic reasoning challenges. Through \textit{DiagnosisArena}, we aim to drive further advancements in AI’s diagnostic reasoning capabilities, enabling more effective solutions for real-world clinical diagnostic challenges. We provide the benchmark and evaluation tools for further research and development.\footnote{\quad\url{https://github.com/SPIRAL-MED/DiagnosisArena}}
\end{abstract}

\vspace{30pt}
\begin{figure}[h]
    \centering
    \scalebox{0.98}{
    \includegraphics[width=\linewidth]{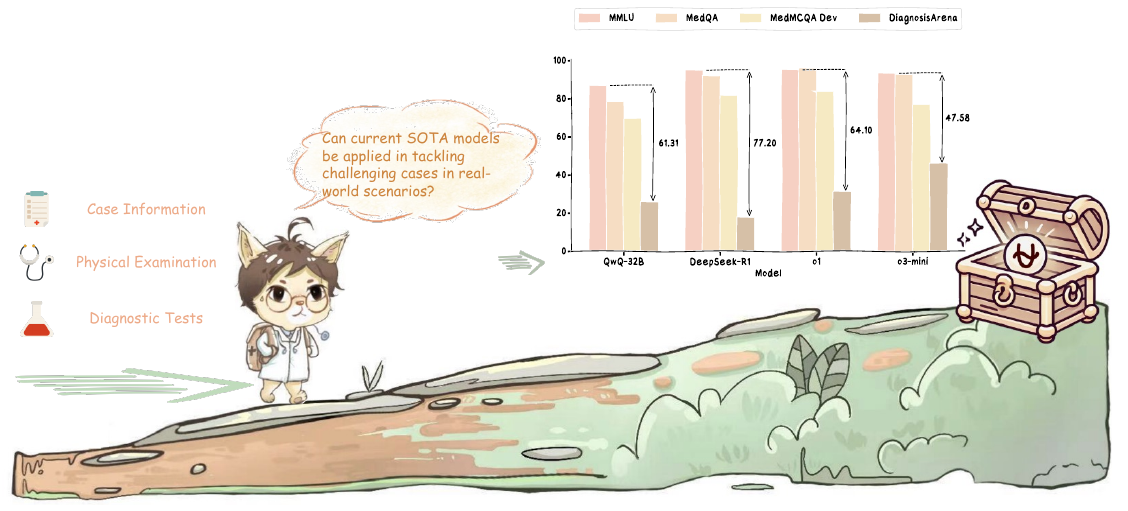}
    }
    \caption{Performance of SOTA models on \textit{DiagnosisArena} and other benchmarks.}
\end{figure}

\newpage

\pagestyle{fancy}
\lhead{}
\renewcommand{\headrulewidth}{0.7pt}
\setlength{\headsep}{5mm}

\section{Introduction}
Recent advances in the reasoning capabilities of large language models (LLMs) have transformed the landscape for addressing complex scientific problems, such as mathematics and programming~\cite{openai2024reasoning, deepseekai2025deepseekr1incentivizingreasoningcapability}. Using step-by-step problem-solving and iterative refinement processes, LLMs have demonstrated performance that exceeds humans in diverse tasks~\cite{qin2024o1, huang2024o1, guan2025rstar, chen2025empirical}. Emerging studies have increasingly highlighted the promising potential of integrating these reasoning capabilities into clinical scenarios~\cite{nori2024medprompt, huang2025o1, chen2024huatuogpt, yu2025finemedlm, jiang2025meds}. However, there remains a considerable gap in assessing the readiness of such models for real-world clinical deployment. Consequently, how to benchmark the clinical diagnostic capabilities of reasoning models has become a focal point for further application.

Existing benchmarks primarily use medical examination-style questions to evaluate the capabilities of LLMs in the medical domain~\cite{nori2024medprompt}. Such tasks are mainly reliant on specific knowledge and have become relatively straightforward for contemporary LLMs. In fact, state-of-the-art models now achieve more precision than 90\% on established benchmarks, including MMLU~\cite{hendrycks2020measuring} and MedQA~\cite{jin2021disease}. The saturation of these existing benchmarks impedes an accurate evaluation of LLM capabilities in realistic diagnostic scenarios. 
Although recent efforts have attempted to construct benchmarks based on real-world clinical cases~\cite{chen2024benchmarking, wang2023cmb, rarearena}, these evaluations often suffer from limitations such as restrictive multiple-choice formats or insufficient clinical relevance. Fundamentally, diagnostic reasoning requires clinicians to identify a specific disease or condition from a set of possible alternatives by analyzing a patient’s symptoms, medical history, physical examination, and diagnostic test results. However, multiple-choice formats inherently constrain the scope of differential diagnostics, artificially simplifying the reasoning process. Moreover, effective clinical decision-making demands that clinicians determine individual patient nuances from extensive and complex information. Existing studies typically perform evaluations using overly simplified case information, thus diverging considerably from real-world clinical complexity. These limitations highlight the urgent need to develop benchmarks capable of rigorously evaluating LLMs in a manner that not only closely mirrors real-world clinical complexities but also emphasizes highly challenging diagnostic cases.

In this paper, we introduce \textit{DiagnosisArena}, a comprehensive and highly challenging benchmark comprising professional diagnostic problems. Given that, in real-world scenarios, treatment recommendations from AI models require extensive validation through prolonged clinical trials and experiments—currently difficult to achieve—we focus primarily on the diagnostic aspect. \textit{DiagnosisArena} is developed through a meticulous construction pipeline, involving data collection, data segmenting, iterative filtering, and expert-AI collaborative verification. Initially, we collected an extensive set of real-world case reports from professional top-tier medical journals—including Lancet, NEJM, JAMA, and so on—to ensure authenticity and diversity. Subsequently, we perform segmented data transformation, converting raw case reports into standardized segmented formats. Each segmented case encompasses detailed \textit{case information}, \textit{physical examination findings}, and \textit{diagnostic test results}, with the final diagnosis serving as the ground truth. To ensure the complexity and quality of the benchmark, iterative filtering was conducted based on analyses from AI experts, coupled with AI-based reviews to verify that each included case contains sufficient and unambiguous information necessary for arriving at the final diagnosis. Moreover, to enhance robustness and minimize potential errors, we employ an expert-AI collaborative verification process: cases are excluded if a consensus on a relevant diagnosis cannot be reached from eight sampled frontier LLMs, or if they fail approval by board-certified physicians. The inclusion of highly challenging cases and a multi-stage curation process culminates in a professional-grade benchmark comprising 1,113 structured clinical cases across 28 medical specialties, designed specifically for evaluating the diagnostic reasoning capabilities of LLMs in complex clinical scenarios.

To quantitatively evaluate the performance of diagnostic outputs, we adopt GPT-4o~\cite{openai2024gpt4o}, one of the most powerful models for knowledge-intensive tasks, as a judge to categorize the relationship between a model's diagnostic results and the ground truth diagnosis as either ``identical'', ``relevant'', or ``irrelevant''. For each case, we generate five candidate diagnostic outputs and calculate both top-1 and top-5 accuracy scores. We conduct extensive experiments involving proprietary models~\cite{openai2024reasoning, openai2024gpt4o, claude-3-5-sonnet, qwen25, openaio3mini, gemini2.5, openai2025o3} and open-source models~\cite{deepseekai2025deepseekr1incentivizingreasoningcapability, baichuan-m1-2025, qwq32b, deepseekai2024deepseekv3technicalreport, qwen3}, including advanced reasoning models (e.g. o1 and DeepSeek-R1). 
Results on \textit{DiagnosisArena} reveal that:
(1) Current leading models exhibit notably low performance in diagnosing professional-level clinical cases. Even the most advanced reasoning models, o3, o1, and DeepSeek-R1, achieve only 51.12\%, 31.09\%, and 17.79\%, respectively, while other models struggle to surpass 20\%. These findings underscore a substantial gap between existing LLM capabilities and the performance required for professional-level medical diagnostics. 
(2) We examine potential data leakage effects by examining the publication dates of the original case reports. Our analysis indicates that instances of data leakage within our benchmark are exceedingly rare, as indicated by the absence of significant divergence in model performance for cases published before versus after the training data cut-off date. This observation highlights the necessity of developing more sophisticated paradigms to enhance AI capabilities for addressing challenging clinical problems.
(3) By converting the diagnostic task in \textit{DiagnosisArena} into the multiple-choice format based on model-generated diagnoses as \textit{DiagnosisArena-MCQ}, we observe a marked increase in model performance, with o1 reaching 61.90\%, further suggesting that multiple-choice formats inherently reduce task difficulty and thus fail to accurately reflect the models' true abilities in addressing complex clinical problems. 

As AI systems approach human-expert performance across various domains, precisely measuring their capabilities and limitations becomes critical for safely integrating these systems into clinical diagnostics. We hope that \textit{DiagnosisArena} will contribute to the ongoing efforts of understanding AI's strengths and weaknesses within the medical domain.

\begin{figure}[t]
    \centering
    \scalebox{1}{
    \includegraphics[width=\linewidth]{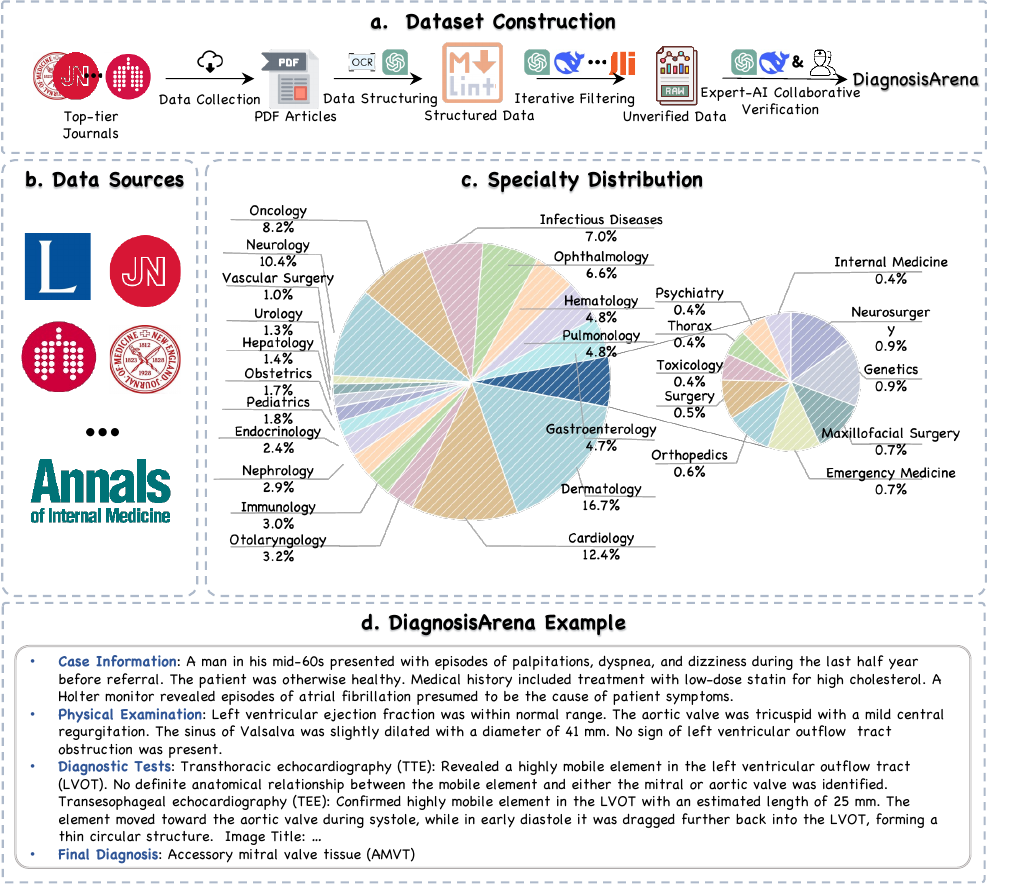}
    }
    \caption{\textbf{Overview of the \textit{DiagnosisArena} Benchmark.} 
    \textbf{(a) }The pipeline for constructing the DiagnosisArena dataset consists of four stages: data collection from the journals, data structuring, iterative filtering of non-reasoning examples, and expert-AI collaborative verification.
    \textbf{(b) }DiagnosisArena is sourced from 10 top-tier medical journals.
    \textbf{(c) }DiagnosisArena is highly diverse, covering 28 medical specialties.
    \textbf{(d) }DiagnosisArena boasts clearly defined segments and offers information-dense clinical cases, which align more closely with clinical practice and present greater reasoning complexity.
    }
    \label{fig: main_fig}
\end{figure}

\section{Related Work}

\textbf{Reasoning LLMs.} 
Endowing LLMs with reasoning abilities has been regarded as a challenging task. As the overall capabilities of LLMs continue to improve, enhancing their reasoning abilities has increasingly become a focal point of research.
Initially, the focus shifted from the few-shot approach to making LLMs mimic the reasoning process to fixed paradigms such as Chain-of-Thought~\cite{wei2022chain} and ReAct~\cite{yao2023react} to stimulate LLMs' reasoning abilities. Later, researchers recognized that reasoning is a process that is gradually optimized through trial, error, and correction. Based on this understanding, heuristic search and process-level reward models were introduced. Heuristic search drew inspiration from traditional search algorithms, such as Monte Carlo Tree Search~\cite{silver2017mastering}, and expanded the LLM’s single chain of thought into a more complex tree structure of thought, as seen in~\cite{yao2023tree, shinn2023reflexion}. On the other hand, the process-level reward model models the reasoning process as a Markov Decision Process, gradually assigning rewards to intermediate steps, guiding the model to generate more accurate chains of thought~\cite{lightman2023let, jiao2024learning, wang2023math}.
Then, the advent of OpenAI's o1~\cite{openai2024reasoning} sparked a wave of research on reasoning. Journey Learning~\cite{qin2024o1} explores multiple strategies to replicate the o1-like slow-thinking reasoning. O1-coder~\cite{zhang2024o1} integrates Monte Carlo Tree Search in code-related domains. STILL-2~\cite{min2024imitate} distills long-form reasoning data, expands potential solution paths, and iteratively optimizes. These efforts culminated in breakthroughs exemplified by DeepSeek's achievements~\cite{deepseekai2025deepseekr1incentivizingreasoningcapability}, which underscore the critical role of reinforcement learning in augmenting reasoning performance. The development of reasoning models is still ongoing...


\textbf{Benchmark for Medical LLMs.}As LLMs continue to advance in medicine, the benchmark for medical applications is also evolving. Before the emergence of LLMs, most commonly medical benchmarks like MedQA~\cite{jin2021disease} and MedMCQA~\cite{pal2022medmcqa} were derived from medical licensing examinations, such as USMLE, CHMLE, or others. They primarily focus on assessing LLMs' mastery of standardized medical knowledge.
As the application scenarios of LLMs expand, benchmarks based on subdomain applications have emerged, such as medical calculators~\cite{khandekar2024medcalc}, medical visual X-ray~\cite{zhou2024benchx}.
However, benchmarks focused on knowledge assessment do not align with expectations for LLMs to perform diagnosis and treatment in clinical settings. Recent studies have increasingly focused on clinical scenarios, with these benchmarks moving away from reliance on medical licensing examinations and instead drawing data from broader sources to better align with real-world clinical scenarios. Examples of such work include Medbullets~\cite{chen2024benchmarking}, CMB-Clin~\cite{wang2023cmb} and RareArena~\cite{rarearena}.
With the development of reasoning models in the medical field, research indicates that traditional benchmarks have gradually lost their challenge~\cite{nori2024medprompt}. As a result, researchers have started to introduce more difficult and multi-level filtration benchmarks~\cite{zuo2025medxpertqa, qiu2025quantifying}. We believe that the fundamental reason for this phenomenon lies in the insufficient assessment of model reasoning abilities in clinical scenarios by traditional benchmarks. We focus on medical diagnosis, a scenario that demands complex reasoning over multidimensional patient records. Building on this, we introduce \textit{DiagnosisArena}, a benchmark designed to comprehensively evaluate diagnostic reasoning.

\section{DiagnosisArena}

\subsection{Overview}
We introduce \textit{DiagnosisArena}, a comprehensive and challenging benchmark for medical diagnosis, designed to evaluate the capabilities of LLMs in diagnosing challenging cases in real-world scenarios.
In Section \ref{sec: construction}, we introduce the construction pipeline, which comprises data collection, data structuring, iterative filtering, and expert-AI collaborative verification.
In Section \ref{sec: comparison}, we compare \textit{DiagnosisArena} with other existing benchmarks. 
With comprehensive case information and alignment with realistic clinical diagnostic scenarios, \textit{DiagnosisArena} serves as an effective benchmark for evaluating the performance of LLMs in addressing complex diagnostic tasks.

\begin{table}[t]
\centering

\resizebox{\linewidth}{!}{
\begin{tabular}{lccccccc}
\toprule 
\textbf{Benchmark} & \textbf{\thead{\# Sample\\Size}} &\textbf{\thead{\# Average\\Length}} & \textbf{\thead{Problem Type}} & \textbf{\thead{Exams \& Tests}} & \textbf{\thead{Clinical\\Scenarios}} & \textbf{\thead{Data Source}} \\
\midrule

\rowcolor{cyan!8} PubMedQA \cite{jin2019pubmedqa} & 1, 000 & 328.41 & \textit{Close-Ended} & \ding{56} / \ding{56} & \ding{56} & PubMed\\
\rowcolor{cyan!8} MMLU (Medical) \cite{hendrycks2020measuring} & 1, 089 & 100.07 & \textit{MCQ} & \ding{56} / \ding{56} & \ding{56} & Licensing Exams\\
\rowcolor{cyan!8} MedQA-USMLE \cite{jin2021disease} & 1, 273 & 215.46 & \textit{MCQ} & \ding{52} /\ding{56} & \ding{52} & Licensing Exams\\
\rowcolor{cyan!8} MedMCQA-Dev \cite{pal2022medmcqa} &  4, 183 & 53.84 & \textit{MCQ} & \ding{56} /\ding{56} & \ding{56} & Licensing Exams \\
\rowcolor{yellow!8} CMExam \cite{liu2023benchmarking} & 6, 811 & 150.67 & \textit{MCQ} & \ding{56} /\ding{56} & \ding{56} & Licensing Exams\\
\rowcolor{yellow!8} C-Eval (Medicine) \cite{huang2023ceval} & 375 & 32.89 & \textit{MCQ} & \ding{56} /\ding{56} & \ding{56} & Licensing Exams\\
\rowcolor{yellow!8} CMB-Clin \cite{wang2023cmb} & 74 & 792.55 & \textit{Open-Ended} & \ding{52} / \ding{52} & \ding{52} & Hospital \\
\rowcolor{yellow!8} MMLU-Pro (Medical) \cite{wang2024mmlu} & 586 & 166.63 & \textit{MCQ} & \ding{56} / \ding{56} & \ding{56} & Licensing Exams\\
\rowcolor{yellow!8} Medbullets \cite{chen2024benchmarking} & 124 & 209.95 & \textit{MCQ} & \ding{52} / \ding{52} & \ding{52} & Question Bank\\
\rowcolor{yellow!8} RareArena~\cite{rarearena} & 72, 661 & 310.36 & \textit{Open-Ended} & \ding{52} / \ding{52} & \ding{52} & PubMed\\
\rowcolor{red!8} MedXpertQA Text\cite{zuo2025medxpertqa} & 2, 450 & 257.43 & \textit{MCQ} & \ding{52} / \ding{52} & \ding{52} & Exams \& Boards \\
\rowcolor{red!8} MedR-Bench\cite{qiu2025quantifying} & 1, 453 & 335.37 & \textit{Open-Ended} & \ding{52} / \ding{52} & \ding{52} & PubMed\\
\midrule
DiagnosisArena & 1,113 & 545.02 & \textit{Open-Ended} \& \textit{MCQ} & \ding{52} / \ding{52} & \ding{52} & Top-tier Journals \\
\bottomrule
\end{tabular}}

\caption{\textbf{Comparisons with existing medical benchmarks.} We categorize existing benchmarks into three types based on chronological milestones: those developed \textcolor{cyan!50}{before the emergence of LLMs}, \textcolor{yellow!60}{after the introduction of LLMs} and \textcolor{red!50}{following the advent of reasoning-based models}. In terms of \textbf{Exams \& Tests}, \textbf{Exams} refer to physical examinations, whereas \textbf{Tests} denote diagnostic tests. And $\textit{MCQ}$ refers to multiple-choice questions. Unlike prior benchmarks that are primarily derived from licensing examinations or focus on simplified cases with limited patient information, \textit{DiagnosisArena} presents significantly greater challenges for state-of-the-art LLMs. This increased difficulty stems from its inclusion of rich patients' records and complex clinical scenarios that cannot be addressed through pretrained knowledge alone.}

\label{table: comparison}
\end{table}

\subsection{Construction}
\label{sec: construction}
In real-world clinical scenarios, diagnostic decision-making requires physicians to meticulously analyze extensive patient data—including medical history, recent behaviors, and laboratory results—to piece together the full picture of the patient's condition. This process inherently demands sophisticated reasoning abilities. Although current reasoning models have demonstrated significant proficiency in medical licensing Exams, their performance in real-world diagnostic contexts remains uncertain. To address this gap, we constructed a benchmark designed to closely replicate authentic clinical situations and to pose substantial reasoning challenges, thereby effectively evaluating the upper limits of existing reasoning models. Our meticulous data development pipeline encompasses data collection, data structuring, iterative filtering, and expert-AI collaborative verification.

\textbf{Data Collection.}
Performing medical diagnosis is an information-dense task, wherein overlooking subtle yet critical observations can lead to entirely different diagnostic conclusions. Therefore, it is essential to ensure sufficient and detailed case information during data collection.
Moreover, since LLMs can perform effectively on trivial diagnostic tasks~\cite{nori2024medprompt}, it is imperative to incorporate more challenging scenarios to better evaluate the upper limits of their capabilities.
Based on these considerations, we focused specifically on case reports published in top-tier, high-impact medical journals, as these reports typically present challenging cases of substantial research value while providing comprehensive diagnostic information. Furthermore, we endeavored to include a broad spectrum of cases spanning various medical specialties to enable our benchmark to thoroughly evaluate the diagnostic proficiency of LLMs across multiple clinical disciplines. Consequently, we conducted an extensive review of numerous medical journals and ultimately selected 10 target journals as our data sources, from which we collected a total of 4,175 case reports.

\textbf{Data Segmenting.}
Since the raw data collected may include treatment details or follow-up information that could imply the actual diagnostic outcomes, it is crucial to distinguish between prognostic and diagnostic information, incorporating only diagnostic-relevant content into our benchmark. To achieve this, we apply a combination of rule-based filtering and model-based segmenting to convert the unsegmented raw data into a standardized Markdown format. Specifically, explicit treatment plans and prognostic information were initially filtered out manually based on chapter headings. Subsequently, we employed \texttt{Claude-3.5-sonnet} to systematically extract and structure the diagnostic-related content. 
The prompts for model-based structuring can be seen in Appendix \ref{apdx: implementation}.
Consequently, each case report was reorganized into four sections: case information, physical examination, diagnostic tests, and final diagnosis. The first three encapsulate data about the patient's clinical presentation, and the last serves as the ground truth.

\textbf{Iterative Filtering.}
Clinical diagnoses must be grounded in both patient-specific information and medical knowledge, guided by inductive reasoning. While certain typical cases can be addressed through straightforward knowledge recall, such scenarios are inadequate for evaluating the true capabilities of LLMs when confronted with complex clinical situations. We perform iterative filtering to ensure the complexity and quality of the benchmark. First, we employ \texttt{Baichuan-M1}, \texttt{DeepSeek-V3}, and \texttt{GPT-4o} to eliminate overly simple cases. Each model conducts two sampling attempts per case. If either model produces a correct answer in any of the two attempts, the case is considered too simple and is excluded. Second, we utilize AI experts to assess whether each remaining case contains sufficient contextual clues to support a logically sound diagnostic pathway. Only cases unanimously judged as reasonable by all AI expert reviewers are retained.
Details of the AI Experts are provided in \ref{apdx: implementation}. 
Following this procedure, we retain 1,783 cases for the next stage.

\textbf{Expert-AI Collaborative Verification.}
To ensure the accuracy of the final diagnostic results and minimize potential errors, we employed an Expert-AI Collaborative Verification mechanism. First, we used the advanced model \texttt{DeepSeek-R1} to perform multiple rounds of sampling and voting. Specifically, cases are excluded if a consensus on a relevant diagnosis cannot be reached in 8 attempts. Next, we enlisted board-certified physicians to conduct reviews. If the experts identified missing information or ambiguity in the diagnosis, the corresponding cases were also excluded. In total, 1,113 cases were excluded through this process.

\subsection{Comparison}
\label{sec: comparison}

\begin{figure}[t!]
    \centering
    \scalebox{1}{
    \includegraphics[width=\linewidth]{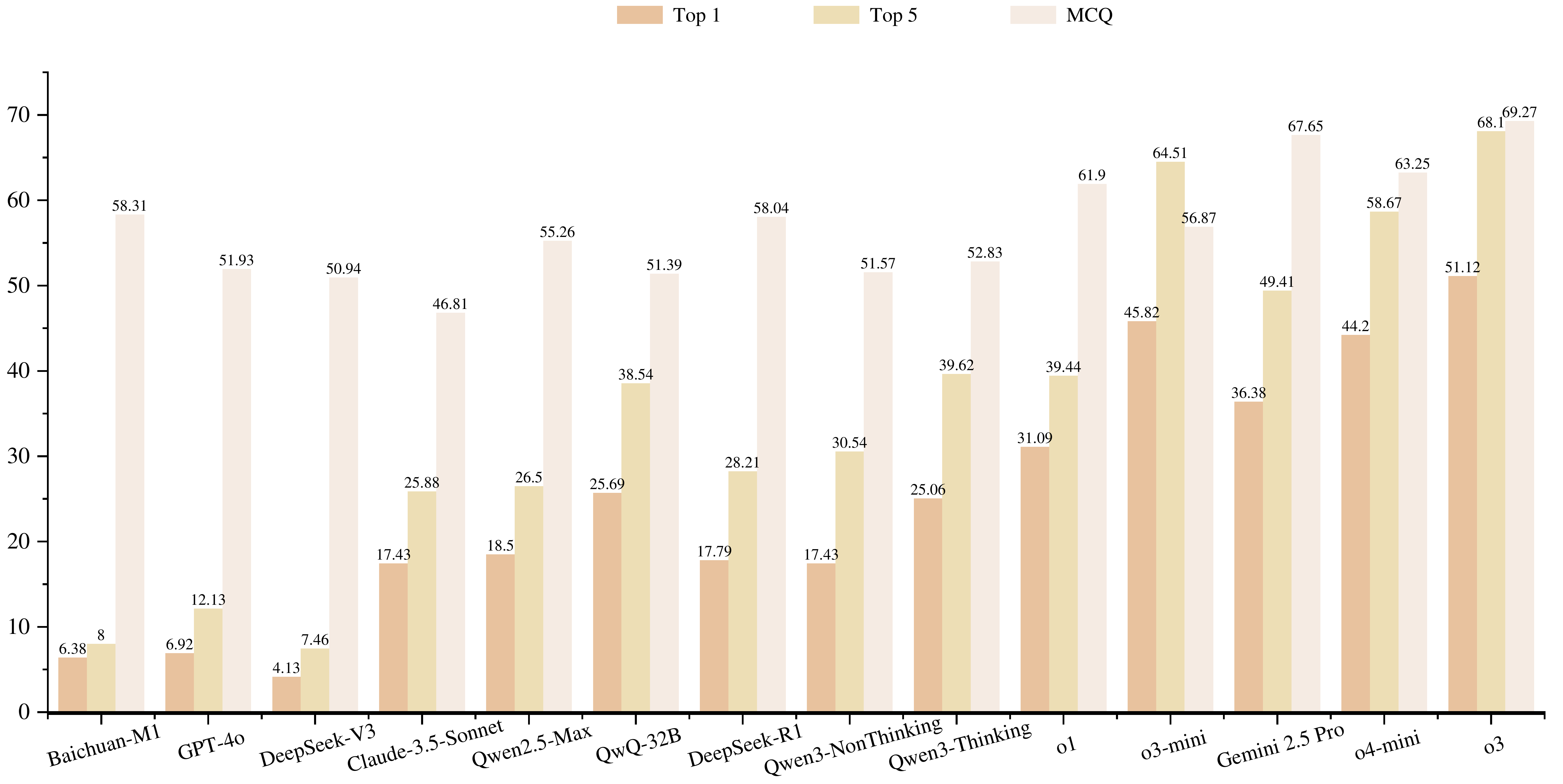}
    }
    \caption{\textbf{Performance of Different Models on \textit{DiagnosisArena}.} (a) The Top-\(k\) metric represents the proportion of cases where the correct answer is included among the Top-\(k\) predictions generated by the model, ranked in descending order of confidence. The results reveal that while the o3 outperforms others, \textit{DiagnosisArena} remains a significant challenge for all existing models. (b) The MCQ presents the multiple-choice version of \textit{DiagnosisArena}. A marked increase in model performance can be observed, with o1 reaching 61.90\%.}
    \label{fig: main_result}
\end{figure}

With the above pipeline, we constructed \textit{DiagnosisArena}, a benchmark comprising 1,113 cases across 28 medical specialties. Table \ref{table: comparison} compares \textit{DiagnosisArena} with other existing benchmarks. We divide existing benchmarks into three categories. The first one consists of traditional and widely used medical benchmarks, with data primarily derived from questions of the Medical Licensing Examination. The second one emerged alongside the development of LLMs, encompassing a wide range of task types—from question banks to clinical case analyses. The third one represents recent benchmarks that appeared following the release of \texttt{OpenAI-o1}. While these benchmarks aim to address more complex medical problems, they often remain limited to multiple-choice formats or lack real-world clinical context.

In contrast, \textit{DiagnosisArena} is sourced from clinical case reports published in 10 top-tier medical journals. These case reports are comprehensive and inherently challenging, making them well-suited for evaluating the diagnostic reasoning capabilities of LLMs. In clinical practice, comprehensive physical examinations and diagnostic tests (e.g., blood tests, CT scans) are essential for assessing a patient’s condition and identifying health issues. By incorporating such clinical information, \textit{DiagnosisArena} closely mirrors real-world medical scenarios and includes detailed patient data such as age, sex, presenting symptoms, examination findings, etc. Furthermore, our rigorous filtering pipeline excludes cases that can be resolved solely using memorized specific knowledge, thereby emphasizing the evaluation of reasoning abilities rather than simple retrieval.

\section{Experiments}

\subsection{Implementation Details}

\textbf{Setup.} To fairly evaluate the diagnostic capabilities of current LLMs on our \textit{DiagnosisArena}, we use a unified prompt to instruct models to generate five possible diagnostic outcomes in descending order of confidence. To minimize potential biases, our prompt did not impose strict constraints on the output format. Details of the prompt are shown in Appendix~\ref{apdx: implementation}. 

\textbf{Models.} Our experiments include both proprietary and open-source models, as well as domain-specific medical models and general models, such as Baichuan-M1~\cite{baichuan-m1-2025}, DeepSeek-V3~\cite{deepseekai2024deepseekv3technicalreport}, GPT-4o~\cite{openai2024gpt4o}, Claude-3.5-Sonnet~\cite{claude-3-5-sonnet}, Qwen2.5-Max~\cite{qwen25}. Notably, we also focused on assessing models that leverage inference-time scaling, such as open-source QwQ-32B~\cite{qwq32b}, OpenAI’s o1~\cite{openai2024reasoning}, DeepSeek-R1~\cite{deepseekai2025deepseekr1incentivizingreasoningcapability}, Qwen3-235B-A22B~\cite{qwen3}, Gemini 2.5 Pro~\cite{gemini2.5}, o3-mini~\cite{openaio3mini}, o4-mini~\cite{openai2025o3} and o3~\cite{openai2025o3}. Additionally, we tested both the thinking and non-thinking configurations of Qwen3-235B-A22B, denoted as Qwen3-Thinking (with enable\_thinking set to true) and Qwen3-NonThinking (with enable\_thinking set to false).

\subsection{Evaluation}

\textbf{Open-Ended Question Evaluation.}
For open-ended questions, the ground truth is a clear diagnostic conclusion. In clinical scenarios, medical diagnoses are generally classified into three categories: ``identical'', ``relevant'', and ``irrelevant''. Therefore, we use \texttt{GPT-4o} as the judge to evaluate the output results into these three categories. Among them, only when the model's result is judged to be ``identical" is it considered correct. Additionally, we instruct the LLM to generate the $k$ possible diagnostic outcomes in descending order of confidence, and then calculate the Top $k$ accuracy, which is the hit rate of the correct answer within the top $k$ predicted results. The prompt is shown in Appendix~\ref{apdx: implementation}.

\textbf{Multi-Choice Question Evaluation.}
After open-ended question evaluations, we select partially correct diagnostic results from the answers generated by LLMs—primarily from the evaluation results of o1 and DeepSeek-R1—as distractor options and construct multiple-choice questions with four options. For multiple-choice questions, the ground truth is a fixed option. We instruct the LLMs to answer and perform rule-based extraction and comparison to calculate the accuracy. The prompt is shown in Appendix~\ref{apdx: implementation}.

\subsection{Main Results}
\begin{figure}[t]
\centering
\subfloat{
\includegraphics[width=0.49\linewidth]{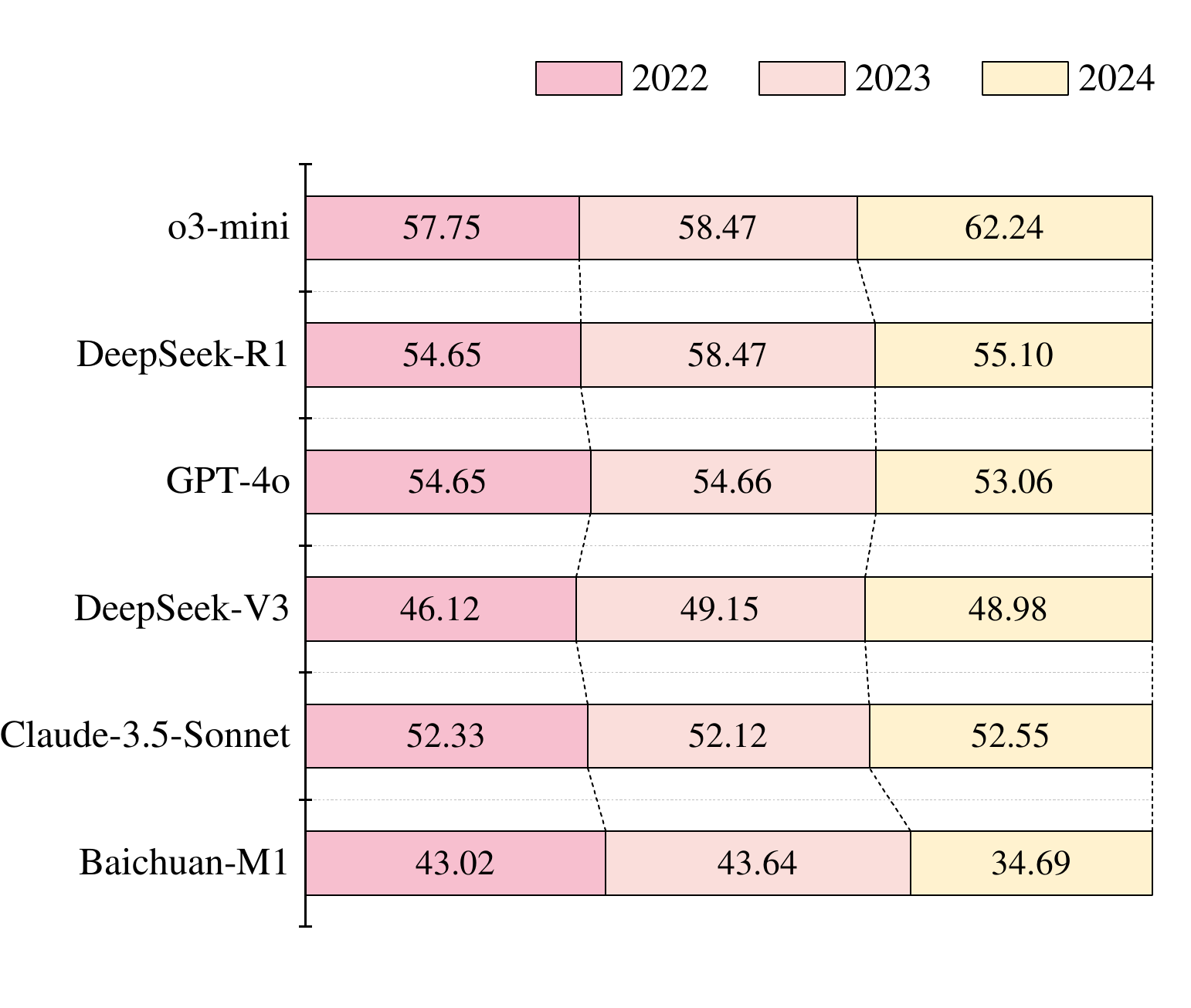}
}%
\subfloat{
\includegraphics[width=0.50\linewidth]{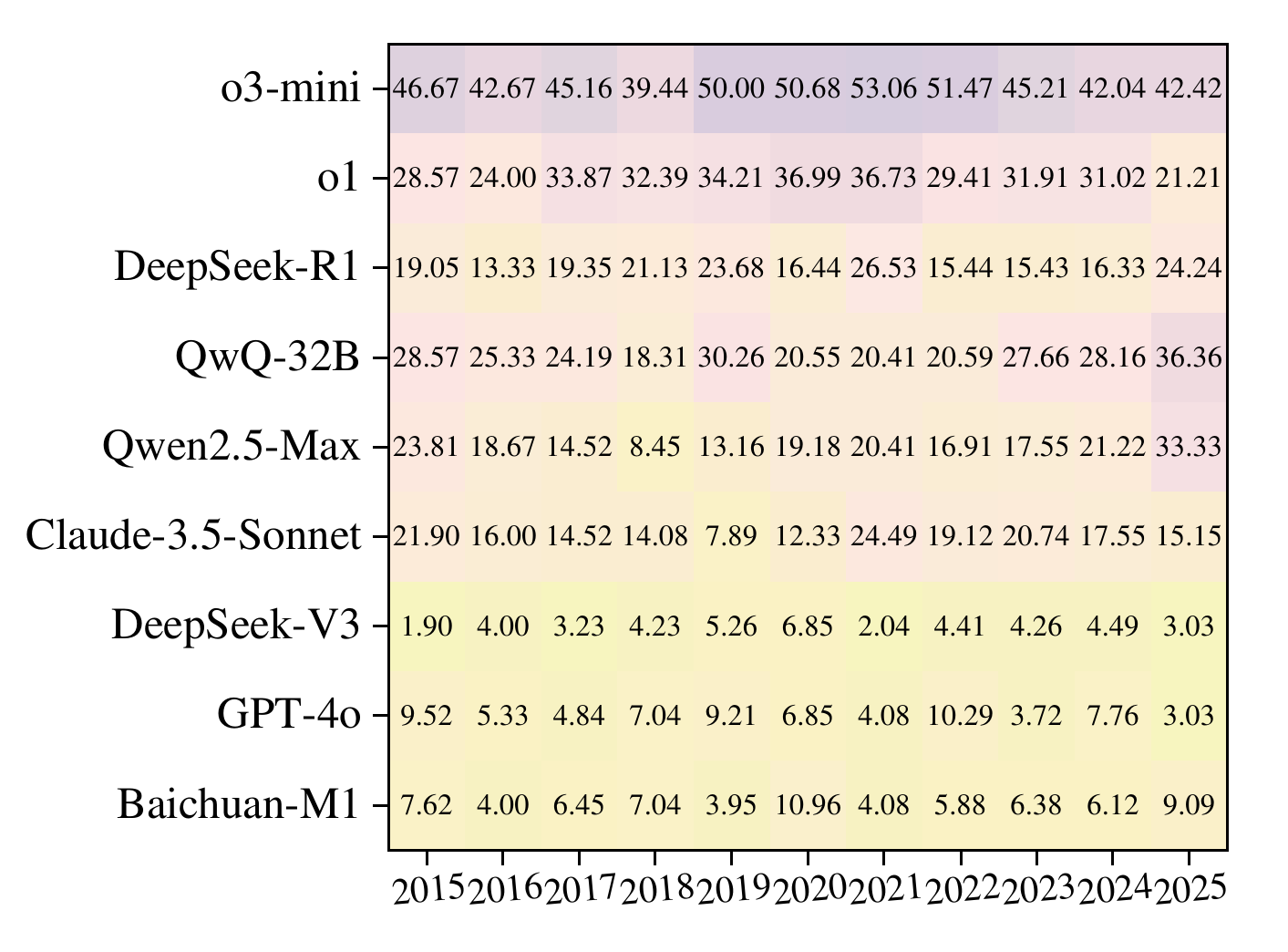}
}%
\caption{\textbf{Leakage Detection on \textit{DiagnosisArena}.} (a) Pre-experiment small sample Leakage Detection. For all models, the experimental results maintained a generally consistent trend across different years, with only minor fluctuations. (b) Leakage Detection on the Constructed \textit{DiagnosisArena}. Over the past decade, all models have demonstrated relatively stable accuracy, with no significant fluctuations over time.}
\label{fig: data_leakage}
\end{figure}

Figure \ref{fig: main_result} presents the main evaluation results of LLMs on the \textit{DiagnosisArena}. Based on our analysis of the experimental data, we draw the following conclusions:
(1) Even the most advanced reasoning LLMs struggle with this task. The best-performing model, o3, achieved an accuracy of only 51.12\% on \textit{DiagnosisArena}, while o1 and DeepSeek-R1 performed even worse, with an accuracy as low as 31.09\% and 17.79\%. This substantial performance gap highlights the significant difficulty of our benchmark and the limitations of current models.
(2) Models endowed with explicit reasoning capabilities demonstrate a clear advantage in clinical diagnostic tasks. Notably, even powerful models such as Claude-3.5-Sonnet and Qwen2.5-Max achieve suboptimal performance—below 20\% accuracy. In contrast, reasoning-enhanced models, including those with relatively smaller parameter sizes (e.g., QwQ-32B), attain significantly higher accuracy, reaching 25.69\%. A particularly illustrative case is DeepSeek-R1, a reasoning-oriented model trained by DeepSeek-V3, which achieves a 13.66\% improvement in accuracy over its base model. These findings underscore the critical role of reasoning in clinical diagnosis.
(3) Solely prompting LLMs to select the best answer in a multiple-choice format does not accurately reflect their true capabilities in handling clinical tasks. A significant performance improvement is often observed in the multiple-choice setting—for example, o1 achieves 61.90\%. Even Baichuan-M1-14B, which scores below 10\% in open-ended settings, attains 58.31\% in the multiple-choice format. This discrepancy can be attributed to the inherently simplified nature of multiple-choice questions, where predefined options narrow the problem space. In such cases, LLMs can leverage superficial cues or rely on partial knowledge to eliminate incorrect choices and select the most plausible answer. However, this does not constitute a complete deductive reasoning process as required in real-world diagnostic scenarios.

\begin{figure}[t!]
    \centering
    \scalebox{1}{
    \includegraphics[width=\linewidth]{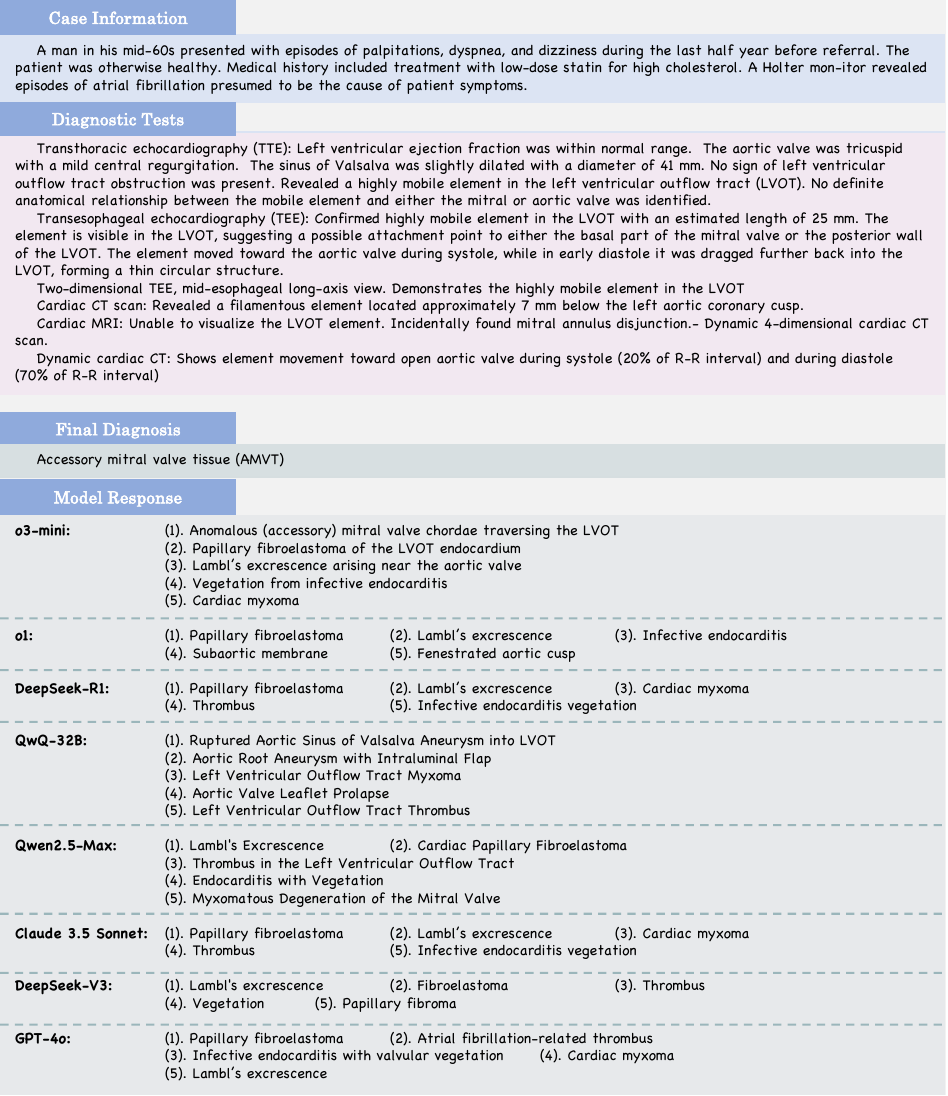}
    }
    \caption{\textbf{A Case Study of \textit{DiagnosisArena}.} Except for \texttt{o3-mini,} which successfully provided the correct answer in the top 1, the other models were far from the correct answer. Analyzing \texttt{DeepSeek-R1}'s response, we found that, despite numerous indirect pieces of evidence supporting the diagnosis of AMVT, \texttt{DeepSeek-R1} selectively ignored these clues and instead overly relied on the reasoning paths of common diseases. \texttt{DeepSeek-R1}'s response is shown in Appendix~\ref{apdx: case}.}
    \label{fig: case_study}
\end{figure}

\subsection{Data Leakage}

The scope of pretraining corpora is increasingly broad, with academic journal papers often included in the pretraining data of LLMs. This may cause LLMs to memorize specific cases and reproduce knowledge to answer questions. To ensure the robustness of \textit{DiagnosisArena}, we conducted experiments to detect data leakage.
To verify the presence of data leakage from sources, we conducted the following analysis.  We conducted the following analysis. We collected 690 journal entries from 2022 to 2024, including publications from JAMA and NEJM. Data from 2025 were excluded due to insufficient volume for a meaningful comparison. The datasets were balanced across years, with approximately 200 entries per year. We performed preliminary processing and evaluation on these datasets, with the results shown in Figure \ref{fig: data_leakage}. 
We can observe that for mainstream models—such as o3-mini, DeepSeek-R1, GPT-4o, DeepSeek-V3, Claude-3.5-Sonnet—remained largely stable across the three years, exhibiting only minor fluctuations. This suggests that either no data leakage occurred or any leakage that did occur had a negligible impact on model evaluation. However, for models in specific like Baichuan-M1, the accuracy exhibited a slight decline in 2024, indicating that data from prior to 2024 may have leaked into its training corpus, thereby affecting its evaluation performance.
Based on the pilot studies, to directly perform data leakage detection on \textit{DiagnosisArena}, we conducted another statistical analysis on the constructed \textit{DiagnosisArena}, with the results shown in Figure \ref{fig: data_leakage}. As observed, compared to the results of the Pre-experiment small sample Leakage Detection, DiagnosisArena exhibited more irregular fluctuations. However, from the overall trend over the decade, neither mainstream general models nor domain-specific medical models showed significant accuracy fluctuations over time. This result indicates that the evaluation in \textit{DiagnosisArena} did not experience performance anomalies due to data leakage.

\subsection{Case Study}
\label{sec: case_study}
In this section, we analyze a case in \textit{DiagnosisArena} and explore the reasons why current reasoning models struggle to solve complex diagnostic problems. 

In this case, as shown in Figure~\ref{fig: case_study}, the final diagnosis is Accessory Mitral Valve Tissue (AMVT). The primary basis for this diagnosis was the discovery of a highly active abnormal structure in the left ventricular outflow tract (LVOT), which exhibited significant motion patterns during the cardiac cycle. The activity and position of this structure, as well as its unclear anatomical relationship with the mitral valve or aortic valve, supported the conclusion that it was an accessory tissue. Various imaging studies and the incidental finding of mitral annulus separation further confirmed the existence of this structure.

Observing the responses of various models, it is evident that, except for \texttt{o3-mini}, which successfully provided the correct answer in the top 1, the other models were far from the correct answer. Analyzing \texttt{DeepSeek-R1}'s response in Appendix \ref{apdx: case}, we can identify the following:
(1) The diagnostic approach did not consider the possibility of AMVT. \texttt{DeepSeek-R1}'s diagnostic reasoning focused on tumors, emboli, fibrolamellar tumors, and other lesions related to the left ventricular outflow tract (LVOT), while overlooking potential mitral valve-related structural issues. 
(2) Misjudgment of imaging features. \texttt{DeepSeek-R1} identified an active structure through imaging and speculated that it might be a papillary fibrolamellar tumor or Lambl's excrescence. However, it did not consider that AMVT might present as a similar filamentous or active structure. 
(3) Inadequate consideration of the correlation with clinical symptoms. \texttt{DeepSeek-R1} linked the patient's symptoms of palpitations, dizziness, and shortness of breath to various diseases such as cardiac tumors, thrombi, and fibrolamellar tumors, but did not sufficiently consider that AMVT might cause mild to moderate left ventricular outflow tract obstruction or intermittent arrhythmias.

We believe that the root cause of this phenomenon lies in the fact that current SOTA reasoning models have not yet fully adapted to the complex reasoning requirements in medical scenarios. Medical reasoning requires attention to every subtle clue, such as the patient's symptoms and the detailed differences in imaging examinations, and gradually piecing together this information to reconstruct the complete truth of the event. However, in our case, despite the existence of numerous indirect pieces of evidence supporting the diagnosis of AMVT, \texttt{DeepSeek-R1} selectively ignored these clues and instead overly relied on the reasoning paths of common diseases. This behavior essentially reflects that the model is still dependent on knowledge reproduction to solve problems, rather than deeply grasping and reasoning through the key details.

\section{Conclusion}
In this paper, we introduce \textit{DiagnosisArena}, a comprehensive and challenging medical benchmark designed to assess the diagnostic reasoning abilities of LLMs in clinical settings. Through a meticulous construction pipeline, \textit{DiagnosisArena} provides 1,113 structured clinical cases across 28 medical specialties, accurately reflecting complex clinical diagnostic scenarios.
The experimental results indicate that even the SOTA reasoning models struggle with \textit{DiagnosisArena}, although they perform relatively well on the multiple-choice version. This significant gap highlights the challenges of diagnostic reasoning in the clinical scenario for current LLMs. It also indicates that simply prompting LLMs to choose the best option from a multiple-choice format does not accurately reflect the models' true reasoning capabilities, which provides a shortcut.
Through the case study, we found that the model tends to prioritize the likelihood of common diseases rather than inferring based on available clues. We conclude that the current models have not yet fully adapted to the complex reasoning demands of clinical diagnostic scenarios.
We hope that \textit{DiagnosisArena} will contribute to advancing reasoning abilities in the field of medicine.

\newpage
\bibliographystyle{acl_natbib}
\bibliography{ref}

\newpage
\appendix

\section{Detailed Statistics of the Benchmark}
\label{apdx: statistics}

\subsection{Distribution of Problems}
We collected 4,175 original data entries from the following ten top-tier journals (shown in Table \ref{table: distribution_journals}). After rigorous screening and validation, we retained 1,113 high-quality data points. These data span 28 medical specialties, with the specific distribution shown in the Figure \ref{fig: main_fig}.

\begin{table}[h!]
\centering
\begin{tabular}{lc}
\toprule
\textbf{Journal Name} & \textbf{Number} \\

\midrule
JAMA & 488 \\
Lancet & 26 \\
Annals of Internal Medicine & 192 \\
Cell & 254 \\
European Respiratory Journal & 18 \\
European Urology & 4 \\
Gastroenterology & 14 \\
Journal of Hepatology & 6 \\
Journal of Thoracic Oncology & 24 \\
NEJM & 87 \\
\bottomrule
\end{tabular}
\caption{The final distribution of Journal Names and their respective Numbers in \textit{DiagnosisArena}.}
\label{table: distribution_journals}
\end{table}

\subsection{Data Leakage Results}
\begin{figure}[ht!]
    \centering
    \scalebox{1}{
    \includegraphics[width=\linewidth]{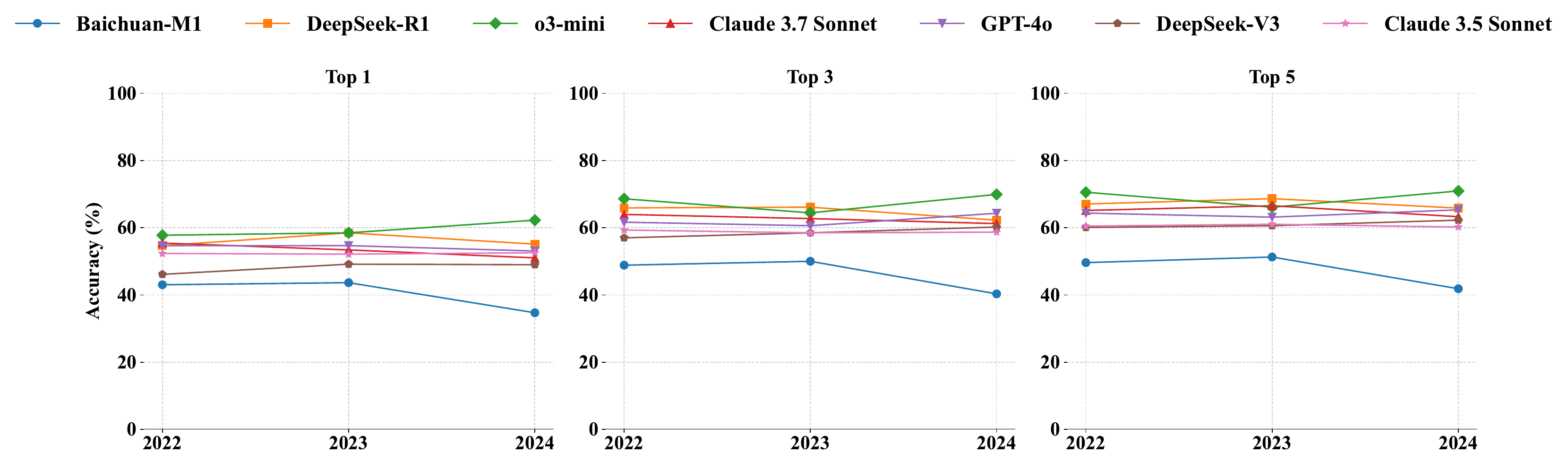}
    }
    \caption{Pre-experiment small sample Leakage Detection.}
    \label{leak1}
\end{figure}

\begin{figure}[ht!]
    \centering
    \scalebox{1}{
    \includegraphics[width=\linewidth]{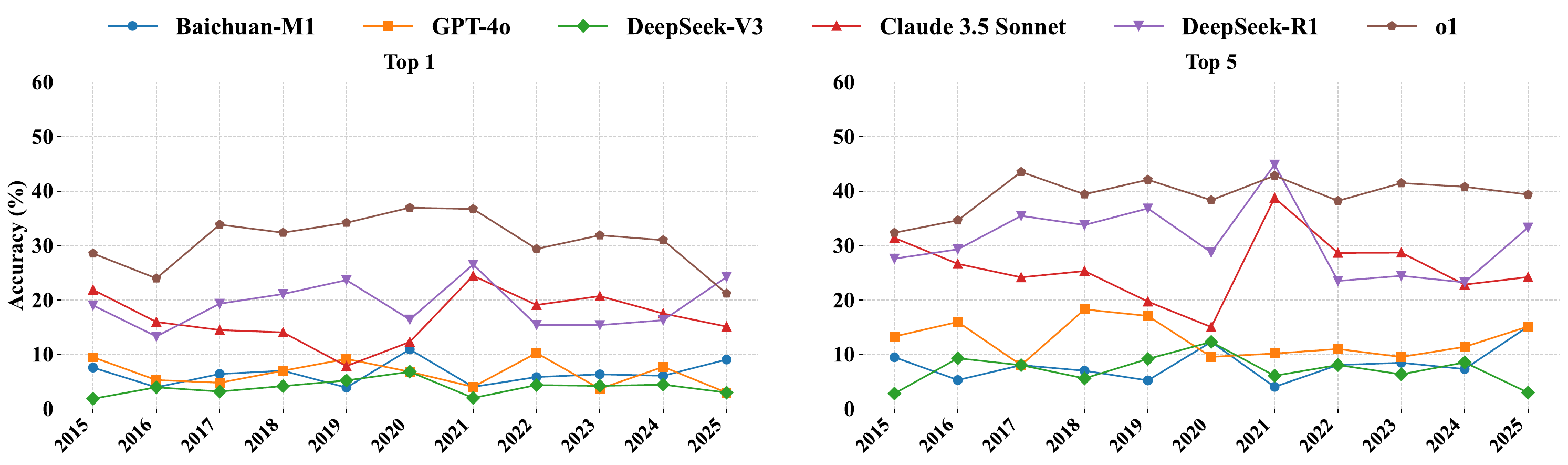}
    }
    \caption{Leakage Detection on \textit{DiagnosisArena}.}
    \label{leak2}
\end{figure}

Comprehensive experimental results regarding the data leakage—encompassing both the primary experiments and the subsequent assessments conducted on \textit{DiagnosisArena}—are illustrated in Figures \ref{leak1} and \ref{leak2}.

\section{Additional Implementation Details}
\label{apdx: implementation}

\subsection{AI Experts}
Specifically, the AI Experts we used include GPT-4o~\cite{openai2024gpt4o}, Claude-3.5-sonnet~\cite{claude-3-5-sonnet}, Qwen2.5-Max~\cite{qwen25}, DeepSeek-V3~\cite{deepseekai2024deepseekv3technicalreport}, DeepSeek-R1~\cite{deepseekai2025deepseekr1incentivizingreasoningcapability}, and o1~\cite{openai2024reasoning}. They are widely used in our data construction pipeline.

\subsection{Prompts}
\definecolor{prompt_color}{RGB}{230, 220, 237}

\begin{tcolorbox}[colback=prompt_color!10!white, colframe=prompt_color!100!white, left=2mm, right=2mm, title=\small\centering\textcolor{black}{\textit{DiagnosisArena} Segmenting Prompt 1}]
Your task is to convert raw text data into clear and well-formatted Markdown files. Please follow the guidelines below:\\

1. Formatting Adjustments  

   \quad - Retain the original paragraph content and apply clear Markdown formatting while maintaining the original layout.  
   
   \quad - Use appropriate Markdown syntax for headings based on the original heading hierarchy (e.g., `\#`, `\#\#`, `\#\#\#`).  \\

2. Removal of Irrelevant Information  

   \quad - Remove all references, citations, footnotes, bibliographies, and any content unrelated to the main topic of the article.  
   
   \quad - Completely delete citations and footnotes, including their in-text markers.\\

3. Handling of Images and Tables  

   \quad - In the main text: Keep all image and table label (e.g., "Figure 1", "Table 1") but do not alter their placement or content.  
   
   \quad - Figure Information Section: At the end of the document, add a new section titled `\#\# Figure Information`, listing all image and table titles and descriptions in Markdown list format. Ensure their numbering matches the original document.  \\

4. Maintaining Content Integrity  

   \quad - Ensure that the actual content remains unchanged, preserving the accuracy and completeness of the original text.  
   
   \quad - Only perform formatting and cleaning adjustments without modifying the original content.  \\

Make sure the final output adheres to Markdown syntax standards, with clear content, neat formatting, and easy readability.
\end{tcolorbox}

\begin{tcolorbox}[colback=prompt_color!10!white, colframe=prompt_color!100!white, left=2mm, right=2mm, title=\small\centering\textcolor{black}{\textit{DiagnosisArena} Segmenting Prompt 2}]
You will receive a medical paper of the type "Case Report." Your task is to **accurately extract** the following four sections and organize them into a JSON-formatted data structure for use in medical exam question design.\\

The definitions of each section are as follows:\\

1. **Case Information**: Includes the patient's basic details (e.g., gender, age, occupation), medical history (past medical history, family history, current medical history), and disease progression.  

   \quad - **Important Requirement**: If the case report explicitly mentions a disease name or diagnostic speculation (e.g., "considering XX disease" or "highly suspected XX"), this information must be removed to avoid directly revealing the final diagnosis.\\

2. **Physical Examination**: Includes the results of the patient's physical examination during the initial consultation or hospital admission, presenting all clinical signs found in the examination. 

   \quad - **Prohibited Content**: Any direct diagnostic statements involving disease names.\\

3. **Diagnostic Tests**: Includes laboratory tests, imaging examinations, pathological examinations, and other auxiliary test results, categorized by test type.  

   \quad - **Handling of Images and Tables**:
     - If the case report contains images or tables of imaging studies, lab reports, etc., extract their **titles** and **descriptions** and classify them under the corresponding test category.  
     
     - Ensure that imaging studies (e.g., X-ray, CT, MRI), laboratory tests (e.g., blood, urine analysis), and pathological tests are **separately categorized** without mixing.  
     
   \quad - **Prohibited Content**: Any direct diagnostic statements involving disease names.\\

4. **Final Diagnosis**: The doctor's final diagnosis for the patient, which should concisely and accurately summarize the disease name or diagnostic conclusion.\\

Output Format Requirements:

\quad - Your response should be a JSON dictionary containing four keys: `"Case Information"`, `"Physical Examination"`, `"Diagnostic Tests"`, and `"Final Diagnosis"`.

\quad - Each key's value should **retain the original wording as much as possible**. If the original text is overly long or unclear, **it may be appropriately condensed for clarity**, but no subjective speculation should be added.

\quad - **Exclude any non-exam-related information**, such as treatment plans, surgical procedures, prognosis, follow-up, etc.\\

Example Output:

\textasciigrave\textasciigrave\textasciigrave json

\{

  \quad "Case Information": "A 45-year-old male patient was admitted due to recurrent chest pain for 3 months. Past medical history includes hypertension for 5 years, and a family history of coronary artery disease.",
  
  \quad "Physical Examination": "Examination: Blood pressure 150/90 mmHg, heart rate 80 bpm, systolic murmur heard at the apex.",
  
  \quad "Diagnostic Tests": "- Laboratory tests: Complete blood count showed no significant abnormalities. Serum biochemistry indicated elevated troponin levels. - Imaging studies: Coronary angiography revealed 70\% stenosis in the left anterior descending artery.  Image Title: Coronary Angiography Results.  Image Description: The angiography shows luminal narrowing in the left anterior descending artery, with an estimated stenosis of 70\%.",
  
  \quad "Final Diagnosis": "Coronary atherosclerotic heart disease."
  
\}

\textasciigrave\textasciigrave\textasciigrave\\

Medical Paper:
\end{tcolorbox}

\begin{tcolorbox}[colback=prompt_color!10!white, colframe=prompt_color!100!white, left=2mm, right=2mm, title=\small\centering\textcolor{black}{\textit{DiagnosisArena} Evaluation Prompt}]

As a medical expert, please make a diagnosis for the patient's disease based on the case information, physical examination, and diagnostic tests. Please enumerate the top 5 most likely diagnoses for the following patient in order, with the most likely disease listed first.\\

Case Information:

\%s\\

Physical Examination:

\%s\\

Diagnostic tests:

\%s\\

Output the diagnosis in numeric order, one per line. For example:

1. Disease A;

2. Disease B;

...
\end{tcolorbox}

\begin{tcolorbox}[colback=prompt_color!10!white, colframe=prompt_color!100!white, left=2mm, right=2mm, title=\small\centering\textcolor{black}{\textit{DiagnosisArena-MCQ} Evaluation Prompt}]

You are an expert in the field of rare diseases. You will receive a medical case file, including Case Information, Physical Examination and Diagnostic Tests. Please conduct a thorough analysis based on the provided information and select the most appropriate diagnosis from the following four options. The final answer must be formatted as $\backslash$boxed\{Correct Option Letter\}.\\

Here is the medical case file: 

Case Information:

\%s\\

Physical Examination:

\%s\\

Diagnostic Tests:

\%s\\

Here are the four options: 

\%s\\

Output Format: Output the final answer in the following format:

Final answer: $\backslash$boxed{Correct Option Letter}
\end{tcolorbox}

\begin{tcolorbox}[colback=prompt_color!10!white, colframe=prompt_color!100!white, left=2mm, right=2mm, title=\small\centering\textcolor{black}{GPT-4o Evaluation Prompt}]

You are an expert in diagnosing challenging cases. You will receive a student's answer containing 5 differential diagnoses, as well as the reference diagnosis. You need to score each diagnosis from the student's answer according to the following rules:\\

2 = The student’s diagnosis exactly matches the reference diagnosis; 

1 = The student’s diagnosis is a broad category that includes the reference diagnosis; 

0 = The student's diagnosis does not meet the criteria for a score of 1 or 2.\\

Here is the student’s answer: 

\%s
\\

Here is the reference diagnosis: 

\%s\\

Output Format: Output the scores in the following format. 

1. Disease 1 name: score X;

2. Disease 2 name: score X;

...
\end{tcolorbox}

\section{Cases}
\label{apdx: case}

\subsection{Benchmark Cases}
In this case, Case Information, Physical Examination, and Diagnostic Tests are components of the patient's medical record. This structured layout is clearer. The Final Diagnosis represents the final diagnostic result based on the patient's medical record. Options and Right Option correspond to the multiple-choice version, \textit{DiagnosisArena-MCQ}, serving as a simplified version of \textit{DiagnosisArena}.
\definecolor{wkorange}{rgb}{1.0, 0.92, 0.80}

\begin{tcolorbox}[colback=wkorange!10!white, colframe=wkorange!100!white, left=2mm, right=2mm, title=\small\centering\textcolor{black}{Case of problems for \textit{DiagnosisArena}}]

\quad\textbf{Case Information:}

\quad\quad\quad A 78-year-old man presented with painful swelling and ulceration of the glans penis of 1 month's duration. His medical history included an elevated white blood cell count and inguinal lymphadenopathy discovered 9 years prior. Four years after initial presentation, he experienced fatigue, weight loss, and loss of appetite. The current illness began 1 month prior to admission when he noticed asymptomatic 'green-yellow welts' and swelling of the glans penis. The swelling became progressively painful and evolved to the size of an 'orange' over 4 days. Previous treatments with topical antibiotics, antifungal agents, hydrogen peroxide soaks, warm compresses, and cephalexin showed no improvement.\\

\quad\textbf{Physical Examination:}

\quad\quad\quad The patient appeared healthy. The glans penis and corona were grossly edematous, with a 3.7 × 3.3-cm tender, firm ulcer on the ventral
surface of the penis with overlying black eschar, adherent yellow crust, and granulation tissue. No urethral discharge was present.\\

\quad\textbf{Diagnostic Tests:}

\begin{itemize}
    \item Laboratory Tests: Routine blood test results were within normal limits. Cultures, assays, and blood tests for viral, bacterial, or fungal etiology were negative.
    
    - Pathology Studies: Biopsy of the glans penis ulcer showed dermal infiltrate of atypical lymphocytes, which were CD20+, CD79a+, and CD5+, within the background of a reactive T-cell infiltrate. PCR analysis revealed a clonal IgH gene rearrangement.
    \item Imaging Studies: PET-CT scan demonstrated FDG-avid abdominal and peripancreatic lymph nodes.
    Image Title: Clinical Appearance of the Penis
    Image Description: Grossly edematous glans penis and corona with a tender, firm ulcer on the ventral surface of the penis with overlying black eschar, adherent yellow crust, and granulation tissue.
    
    Image Title: Histologic Evaluation of Penile Lesion
    Image Description: Hematoxylin-eosin staining showing dermal infiltrate of large, atypical lymphocytes within a reactive T-cell infiltrate (original magnification ×100)
\end{itemize}

\quad\textbf{Final Diagnosis:}

\quad\quad\quad Richter's transformation presenting as a penile ulcer (transformation of chronic lymphocytic leukemia to diffuse large B-cell lymphoma)\\

\quad\textbf{Options:}

\begin{itemize}
    \item A: Chronic Lymphocytic Leukemia/Small Lymphocytic Lymphoma (CLL/SLL)
    \item B: Diffuse Large B-cell Lymphoma (DLBCL)
    \item C: CD5-positive diffuse large B-cell lymphoma
    \item D: Richter's transformation presenting as penile ulcer (transformation of chronic lymphocytic leukemia to diffuse large B-cell lymphoma)
\end{itemize}

\quad\textbf{Right Option:} D

\end{tcolorbox}

\subsection{Case Study R1 Response}

This response corresponds to the patient case in Figure~\ref{fig: case_study}, which is analyzed in detail in Section~\ref{sec: case_study}.

\begin{tcolorbox}[colback=wkorange!10!white, colframe=wkorange!100!white, left=2mm, right=2mm, title=\small\centering\textcolor{black}{R1 Response of the Figure~\ref{fig: case_study} Case}]
$\langle$think$\rangle$

Okay, let's tackle this case. So the patient is a man in his mid-60s with episodes of palpitations, dyspnea, and dizziness over six months. He's on a statin for high cholesterol but otherwise healthy. The Holter monitor showed atrial fibrillation, which they thought was causing the symptoms.

Looking at the physical exam: normal LVEF, tricuspid aortic valve with mild regurgitation, dilated sinus of Valsalva (41 mm), no LVOT obstruction. The diagnostic tests are key here. TTE and TEE show a highly mobile element in the LVOT, 25mm long. It moves toward the aortic valve in systole and forms a thin circular structure in diastole. The cardiac CT found a filamentous element 7mm below the left coronary cusp. MRI didn't see the LVOT element but found mitral annulus disjunction. The dynamic CT shows movement with the aortic valve opening.

So, possible diagnoses. The mobile structure in LVOT is central. First thought is a cardiac tumor, like a myxoma. But myxomas are usually in the atria. Fibroelastoma? Those are small, mobile, and often on valves. But the TEE didn't link it to the valves. Maybe a papillary fibroelastoma? They can be on the LVOT structures.

Another possibility is a Lambl's excrescence. Those are filamentous, usually on valves, but maybe this one is in the LVOT. They can be mobile and cause embolic events. The filamentous description on CT fits Lambl's.

Also, considering mitral annular disjunction, but that's an incidental finding here. The main issue is the LVOT mass. Could it be a thrombus? But in the LVOT without severe LV dysfunction? Less likely since ejection fraction is normal. 

Another thought: a vegetation from endocarditis. But the patient's history doesn't mention fever or other signs of infection. Also, the element is described as filamentous and mobile, which might not fit typical vegetation. 

Then, maybe a flail mitral valve leaflet or chordae, but TEE didn't link it to the mitral valve. The movement described—toward the aortic valve in systole and back in diastole—suggests something attached in the LVOT. 

Cardiac CT showing filamentous structure below the coronary cusp. Maybe a congenital band or a false tendon? But those are usually benign. However, this one is causing symptoms like palpitations and dizziness, possibly due to intermittent obstruction or triggering arrhythmias (like the atrial fib detected). 

Putting it all together: The top differentials would be:

1. Papillary fibroelastoma: common on valves, but can be in LVOT. Highly mobile, filamentous, causing embolic symptoms or arrhythmias. The size (25mm) is a bit large but possible.

2. Lambl's excrescence: smaller, filamentous, typically on valves. Could be in LVOT, leading to similar symptoms.

3. Myxoma: though typically atrial, but rare in LVOT. Mobile and can cause obstruction.

4. Thrombus: less likely given normal LV function, but possible if there's another source.

5. Vegetation: if considering infective endocarditis despite lack of symptoms.

Wait, but the movement pattern and CT findings might make fibroelastoma more likely than Lambl's. Also, the length is 25mm, which is larger than typical Lambl's. Fibroelastomas can be larger. So maybe fibroelastoma is first, then Lambl's. 

Another possibility is a ruptured chordae tendineae, but TEE didn't show mitral valve involvement. Or a flail aortic valve structure, but the aortic valve is tricuspid with mild regurgitation. 

Mitral annular disjunction is incidental here, so not the main issue. 

So top 5:

1. Papillary fibroelastoma

2. Lambl's excrescence

3. Myxoma

4. Thrombus

5. Vegetation

But also, could there be a false tendon or another structure? But those usually don't cause symptoms. The key is the mobile mass in LVOT causing possible embolic events (leading to AFib) or obstruction. Fibroelastoma is the most likely, followed by Lambl's. Myxoma less likely in LVOT. Then thrombus and vegetation.

$\langle$/think$\rangle$

1. Papillary fibroelastoma;

2. Lambl’s excrescence;

3. Cardiac myxoma;

4. Thrombus;

5. Infective endocarditis vegetation
\end{tcolorbox}

\end{document}